\documentclass[twoside,11pt]{article}

\usepackage{obs_study_style}

\ShortHeadings{Revisiting Rashomon}{D'Amour}
\firstpageno{1}

\begin{document}

\title{Revisiting Rashomon: A Comment on ``The Two Cultures''}

\author{\name Alexander D'Amour \email alexdamour@google.com \\
       \addr Google Research\\
       Cambridge, MA, USA 
       }

\maketitle

\begin{abstract}
Here, I provide some reflections on Prof. Leo Breiman's ``The Two Cultures'' paper.
I focus specifically on the phenomenon that Breiman dubbed the ``Rashomon Effect'', describing the situation in which there are many models that satisfy predictive accuracy criteria equally well, but process information in the data in substantially different ways.
This phenomenon can make it difficult to draw conclusions or automate decisions based on a model fit to data.
I make connections to recent work in the Machine Learning literature that explore the implications of this issue, and note that grappling with it can be a fruitful area of collaboration between the algorithmic and data modeling cultures.
\end{abstract}

\section*{Reflecting on Rashomon}
Even twenty years after publication, Prof. Leo Breiman's ``The Two Cultures'' paper \citep{breiman2001statistical} is a thought-provoking read. I have personally found it to be a useful yardstick for self-reflection, and will revisit it from time to time to measure my own intellectual journey in Statistics. For this reason, I am grateful for the opportunity provided by the editors of Observational Studies to comment on the paper on its twentieth anniversary.

In this comment, I will focus on an idea from Prof. Breiman’s paper that I consider a central challenge to incorporating data modeling sensibilities into the algorithmic modeling culture: the Rashomon Effect, or “multiplicity of good models” phenomenon.
This phenomenon refers to the fact that there are often many candidate models that achieve indistinguishably good ``fit" on a distribution of observed data (now usually measured as accuracy on a randomly held-out test set), but achieve that fit by processing the data in substantially different ways.
In short, this phenomenon is analogous to a lack of identification.
I will highlight some recent work that examined the practical implications of the Rashomon Effect (under the name ``underspecification'') in production-like machine learning systems \citep{d2020underspecification}, then discuss how deeper engagement with the subject matter of the problem provides promising ways to address this concern.

I first read ``The Two Cultures'' as a PhD student firmly (perhaps dogmatically) in the camp of data modelers. However, I found through consulting, as Breiman did, that an expanded toolkit was often useful for solving problems in new contexts, particularly when the goal was narrowly defined. Now, I find myself in a “AI (Artificial Intelligence) first” research organization that is firmly committed to the algorithmic modeling culture, where I am often in the position of advocating for approaches more aligned with data modeling once again. This progression has informed my perspective considerably.

One of the key developments in the last twenty years has been the explosion of the algorithmic modeling community, both in size and in relevance.
The algorithmic modeling culture has made amazing concrete progress over the last decade on previously intractable problems, to the point that many of these advances are now central components of products that shape many of our everyday lives.
Along with this success has come a drive to improve algorithmic models on dimensions that have been championed by the data modeling culture.
To paraphrase Prof. Cox in his comment \citep{cox2001statistical}, these considerations include: linkage with background knowledge, granting insight into a generative or causal process, interpretability, and propagating uncertainty from well-defined sources of noise.
As these issues become central to discussions about the limitations of machine learning and AI, I see opportunities for people who identify strongly with the data modeling camp to make major contributions.

\section*{Rashomon, Ramped Up}

As Breiman describes it: ``What I call the Rashomon Effect is that there is often a multitude of different descriptions [equations $f(\mathbf{x})$] in a class of functions giving about the same minimum error rate.'' 
This effect is particularly interesting when these descriptions are ``distant in terms of the form of the model.''
In a line of work inspired by this idea \citet{fisher2019all} dub a set of models with this property a ``Rashomon set''.
I came to understand the implications of this phenomenon through an inexact analogy to identification in causal inference: in the same way that are many causal models that are compatible with a given probability distribution over observables, a Rashomon set instantiates many functions that predict indistinguishably well in a given prediction problem. 
Breiman presents the Rashomon Effect as an argument against drawing conclusions from simple models that fit the data well, an argument that is extended by \citep{marx2019predictive}.

However, if the Rashomon Effect is a problem for simple model families, it is even more pronounced in families with many more degrees of freedom.
One might imagine that the Rashomon problem could diminish as models become ever more accurate in their predictions on held-out data; after all, emulating the behavior of ``nature's black box'', as Breiman put it, might also constrain a model to replicate nature's inner workings.
What we have seen in the last decade has been the opposite.
Predictive gains in fields like computer vision and natural language processing have been achieved by searching ever-more-complex model spaces, and the most successful model specifications are overparameterized parametric neural network models.
Within those overparameterized families, the models that our training algorithms return often ``solve'' the prediction task in bizarre ways.
For example, in computer vision, work on adversarial examples and texture bias have shown that neural network models often make use of ``shortcut'' features in images that are either invisible to people or secondary to the features most people use to recognize an object \citep{DBLP:journals/corr/GoodfellowSS14,ilyas2019adversarial_neurips,geirhos2018imagenettrained,geirhos2020shortcut}.
As it turns out, there are many more ways to predict well on a test set than we initially imagined.

Breiman's Rashomon Effect has practical implications, even in contexts where the purpose of the model is to automate a process rather than to gain scientific insight, for example, in applications like machine translation, clinical risk prediction, or credit scoring, which Dr. Hoadley discusses in his comment \citep{hoadley2001statistical}.
In these cases, while strong predictive performance on the distribution that generated the training data is typically a necessary condition for a model to be useful, it is often not sufficient.
For example, we may also require that the model exhibit robustness to data corruption or differences in measurement instruments; invariance to socially sensitive features; or usage of information that accords with the knowledge of end users or experts.
Because these properties are not directly constrained by a predictive criterion, a model’s behavior on these dimensions can be rather unpredictable, even between models with identical predictive performance.

In recent work, 39 colleagues and I conducted a large-scale study of Rashomon effects in production-scale machine learning systems \citep{d2020underspecification}.
We dubbed the problem underspecification.
The systems we examined included image classifiers, medical imaging systems, natural language processing systems, and clinical risk prediction systems, based on neural networks.
The general design was as follows.
First, for each model, we developed a metric that probed a practically important dimension of the model’s behavior that was not directly measured by predictive performance on the test set.
We called these probes “stress tests”, and examples included the model's accuracy on corrupted inputs (e.g., images that had been pixelated), the stability of the model’s predictions across irrelevant perturbations (e.g., the gender of pronouns, or the time of day in a medical record), etc.
We then trained several subtly different versions of the model (usually, the models only differed by their random weight initialization at the start of training) on the same data using the same heuristics, and checked that the final collection of models had largely the same performance on the test set.
Finally, we evaluated each of these model replicates on the stress test.
Across each Rashomon set of models, despite having near-identical test set performance, we observed substantially higher variability (sometimes an order of magnitude more) on stress tests.
For example, we found that the sensitivity of image classifiers to pixelation, the sensitivity of language models to gender information, the sensitivity of medical imaging models to camera type, and the sensitivity of clinical risk predictors to physician behavior were all affected in substantial ways by the choice of random seed.
Our conclusion was that these practically important dimensions of model performance are not well-constrainted by standard predictive criteria, and are often determined by arbitrary (even random) choices made during training.
Given the sheer number of ``knobs to twiddle'' in modern machine learning models (a phrase Prof. Efron's comment \citep{efron2001statistical}), we conjectured that much larger Rashomon sets are likely lurking in many successful applications of machine learning.
I only realized after we completed the project that Breiman had included a small version of this experiment in Section 8!

The Rashomon effect has also been used (far more delicately) to study other properties of machine learning models that remain underspecified by predictive performance.
In this vein, \citet{fisher2019all} proposed notions of variable importance as functions of Rashomon sets and \citet{semenova2019study} studied a number of implications of the size of Rashomon sets within model families.
 
\section*{Responses to Rashomon}

To be clear, the Rashomon effect is not a reason to abandon machine learning or other algorithmic approaches.
On the contrary, especially for problems with complex, unstructured data, algorithmic approaches are still some of the best tools we have for understanding and exploiting associations.

Breiman’s suggested (and, as he admitted, incomplete) approach was to simply average some of the variability in a Rashomon set away, which is a common approach today.
This is helpful for stability, and does often lead to improvements on standard predictive performance metrics; for this reason, it is also a common practice in deep learning \citep{Lakshminarayanan2017deepensembles}.
However, in my opinion, this still leaves fundamental issues unaddressed.
After all, when we encounter a causal problem where the estimand is not identified, averaging together contradictory causal models that are equally compatible with the observed data rarely yields the right answer.
We show in the appendix of our study that averaging together different models can lead to more favorable behavior on stress tests, but in cases where the variability in stress test performance is high, the averaged model can perform worse on stress tests than the best individual model in the ensemble.

In my opinion, the Rashomon phenomenon can only be resolved through deeper engagement with how the problem is set up, encompassing issues that Breiman, Efron, and Cox all raised in this discussion.
These include more precise specifications of requirements, which could come in the form of stress tests, or more clearly defined estimands; better incorporation of domain knowledge, in the form of well-motivated constraints or incorporation of auxiliary variables; and, of course, better design in data collection.
One of the key costs of working with algorithmic models is that they come with few inherent guarantees, which means that more processes need to be built around them to explicitly check that they are behaving as expected and making use of the right information.
Here, there are opportunities for those who have honed their skills in judgement around data analysis---no matter their allegiance to data modeling or algorithmic modeling---to make a difference.

\bibliography{DAmour}

\end{document}